\documentclass[]{interact}

\usepackage{color, soul, framed}
\usepackage{epstopdf}
\usepackage[caption=false]{subfig}
\usepackage{cite}
\usepackage{amsmath,amssymb,amsfonts}
\usepackage{algorithmic}
\usepackage{graphicx}
\usepackage{textcomp}
\usepackage{wrapfig}
\usepackage{amsmath}
\usepackage{siunitx}
\usepackage{color}
\usepackage{ulem}
\usepackage{bm}      
\usepackage{parskip}
\usepackage{indentfirst}
\usepackage{amsfonts,amssymb}
\usepackage{float}
\usepackage[numbers,sort&compress]{natbib}
\bibpunct[, ]{[}{]}{,}{n}{,}{,}
\renewcommand\bibfont{\fontsize{10}{12}\selectfont}
\makeatletter
\def\NAT@def@citea{\def\@citea{\NAT@separator}}
\makeatother

\theoremstyle{plain}

\theoremstyle{definition}

\theoremstyle{remark}

\begin{document}

\articletype{ARTICLE TEMPLATE}

\title{Adaptive Finite-Time Model Estimation and Control for Manipulator Visual Servoing using Sliding Mode Control and Neural Networks}

\author{
\name{Haibin Zeng\textsuperscript{a}\thanks{CONTACT Haibin Zeng. Email: zenghaibin{\_}hit@163.com}, Yueyong Lyu\textsuperscript{a}\thanks{CONTACT CORRESPONDING AUTHOR. Email: lvyy@hit.edu.cn}, Jiaming Qi\textsuperscript{a}, Shuangquan Zou\textsuperscript{a}, Tanghao Qin\textsuperscript{a}, and Wenyu Qin\textsuperscript{a}
}
\affil{\textsuperscript{a}School of Astronautics, Harbin Institute of Technology, Harbin, People’s Republic of China}
}

\maketitle

\begin{abstract}
The image-based visual servoing without models of system is challenging since it is hard to fetch an accurate estimation of hand-eye relationship via merely visual measurement. Whereas, the accuracy of estimated hand-eye relationship expressed in local linear format with Jacobian matrix is important to whole system's performance. In this article, we proposed a finite-time controller as well as a Jacobian matrix estimator in a combination of online and offline way. The local linear formulation is formulated first. Then, we use a combination of online and offline method to boost the estimation of the highly coupled and nonlinear hand-eye relationship with data collected via depth camera. A neural network (NN) is pre-trained to give a relative reasonable initial estimation of Jacobian matrix. Then, an online updating method is carried out to modify the offline trained NN for a more accurate estimation. Moreover, sliding mode control algorithm is introduced to realize a finite-time controller. Compared with previous methods, our algorithm possesses better convergence speed. The proposed estimator possesses excellent performance in the accuracy of initial estimation and powerful tracking capabilities for time-varying estimation for Jacobian matrix compared with other data-driven estimators. The proposed scheme acquires the combination of neural network and finite-time control effect which drives a faster convergence speed compared with the exponentially converge ones. Another main feature of our algorithm is that the state signals in system is proved to be semi-global practical finite-time stable. Several experiments are carried out to validate proposed algorithm’s performance. 
\end{abstract}


\section{Introduction}

\noindent The manipulator visual servoing related research is attractive because it usually involves a highly nonlinear system that covers many fields, such as image processing\cite{1}, neural network\cite{2}, and dynamic control\cite{3}, etc. Visual servoing is not only a challenging problem in the research field, but also has broad application prospects in the practical engineering field. The challenging problem has revealed its potential in many economic tasks, like building inspection\cite{4}, assembling with robot arm\cite{5}, grasping missions with manipulators\cite{6}, etc. Visual servoing control methods are roughly classified into position-based visual servoing and image-based visual servoing\cite{7,8}. The former uses the camera model to calculate the position and posture of the target from the feedback image information. Then it uses the inverse kinematics model of the robotic arm to calculate the joint space control law and finally drives the robotic arm’s end to reach the desired position\cite{9}. This method is clear and concise, and the control method can achieve high accuracy if the mathematical model of the system is accurate\cite{10}. However, the camera need to be calibrated periodically because the method has high requirements for the model’s accuracy\cite{11}. Furthermore, this method has high requirements for the working environment because the precision instruments are susceptible\cite{12,13}. Considering that the application sites may be in a harsh environment with high temperature and heavy smoke, and the frequent calibration of the fragile depth cameras costs a lot of labor and financial resources, the position-based vision servoing has its shortcomings for economic interests\cite{14,15}.

Due to the apparent shortcomings of position-based visual servoing mentioned above, many researchers dig into image-based visual servoing. Compared with the position-based visual servoing, the image-based visual servoing directly uses the difference between the desired position and the current position information captured by the camera to obtain the image feature error and directly uses it to calculate the control law of the joint thus achieving a closed-loop system\cite{16}. The control method does not involve the mathematical model of the camera and the robot arm, so this method requires less accuracy of the model and dramatically reduces the workload of the maintenance of the relevant instruments.

The image-based uncalibrated visual servoing (IBUVS) is challenging because the accurate estimation of the hand-eye relationship is important to the system’s convergence performance and robustness against the variant environment\cite{17}. The relationship between the velocity of joint space and image feature can be expressed with a Jacobian matrix. The Jacobian matrix reveals the local hand-eye relationship, and one critical section of IBUVS is to estimate the Jacobian matrix in real-time. Many studies estimate the Jacobian matrix in a purely online or offline way. In a purely online method, the Jacobian matrix is estimated with an algorithm like Kalman filter\cite{18}. The purely online method is a kind of data-driven algorithm, where only input and output data are used to realize the estimation and system control, and no more information about the system is involved\cite{19}. For these reasons, the online method has better robustness against the time-variant factors. Whereas system convergency performance may be less than optimal due to less data on the system model. In a purely offline method, the Jacobian matrix is estimated with a fixed model (neural network, for example) trained with data collected from the system. When only the offline method is used, no real-time updating approach is involved. The system robustness may not be as good as expected because the offline trained algorithm is formed with data collected in a particular situation\cite{20}. It means that the system can hardly cope with the emergencies, which is not considered in controller design but may happen in practical manipulation. However, suppose the difference of the system model between designing and manipulating processes are within a relatively reasonable range, the offline method can possess excellent convergence ability due to a large amount of input and output data in the designing process.

\begin{figure*}[tbp]
	\centering
	\includegraphics[width=14cm]{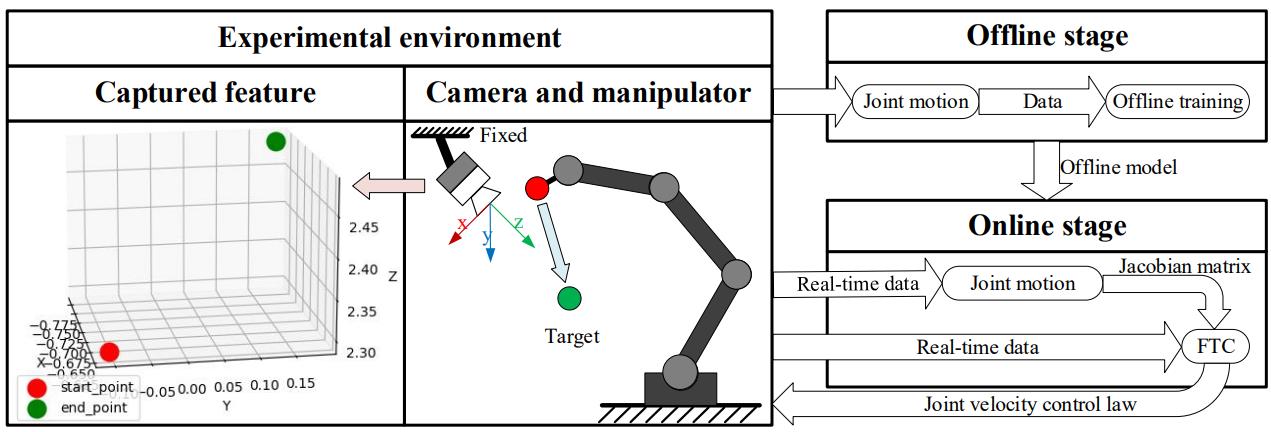}
	\caption{Representation of visual servoing for end-effector position. A red ball is attached to the end of manipulator to make image recognition easier.}
\end{figure*}

To obtain a balance between convergence performance and system robustness against changes in environment and parameters, a Jacobian matrix estimator with both offline and online methods is used in this paper. To begin with, offline training is carried out with data collected from the camera in a particular situation to provide a relatively precise initial estimation for the system. Then an adaptive online update algorithm is executed with real-time input and output data, thus adjusting the offline trained model. Therefore, the system can resist the changes in the model and environment.

Another focus of this article is the system’s working time, that is, finite-time control (FTC). FTC is widely applied for the demands of rapid and accurate manipulator control\cite{21}. The main characteristic of FTC is to ensure that the system state variables converge to the equilibrium or the small neighborhood of the equilibrium in finite time\cite{22}. The finite-time control method is primitively constructed with terminal sliding mode control. Driven by its broad application prospects, many finite-time controllers for nonlinear systems were developed with Lyapunov theory.

Inspired by the above observations, an adaptive finite-time controller based on a neural network is proposed. This paper focuses on the position control of the UR5 robot arm end-effector. The algorithm used in this paper is divided into two sections. The first section is the primitive offline training section. In the first section, Radial Basis Function Neural Network (RFBNN) is used to estimate the Jacobian matrix. The training data set for RBFNN are collected in advance by random joint space motion of UR5, that is, the training set is built under certain premises. Of course, the offline trained RBFNN is not suitable for all situations. The second section is the online update section. When it comes to the manipulation, a controller as well as an update algorithm based on Semi-Global Practical Finite-time Stable (SGPFS) are proposed, where the input and output data of the system are measured in real-time to support the online update program which improves parameters of RBFNN to ensure the estimation for Jacobian matrix can keep pace with the true value. The proposed estimator is later proved to outperform other data-driven estimators with several experiments, which shows that the structure of the combination of online and offline methods is better than the totally online ones. Given the estimated Jacobian matrix, a finite-time controller is designed with a Lyapunov function consisting of both system error and estimation error, which is theoretically better than the scheme that only considers system error and holds the assumption that the estimator is reliable and accurate. What’s more, adaptive factor is introduced in controller designing which help the system converge in fast speed when system error is large and help the system stable when the system error is small. By using this two-step scheme, the robustness of the system is also improved while ensuring the convergence performance of the system.

\section{Preliminaries}
\noindent In this paper, we use standard notation. Lowercase bold letters $\bm{x}$ indicate column vectors. And uppercase bold letters $\bm{X}$ indicate matrixes. Time-variant variables are denoted as
$\bm{x}_{k}$, whose subscript indicates the number of discrete time moment.

In this paper, we focus on the visual servoing for the manipulator’s end-effector without a specific model of the camera or robot. The control effect is realized via the real-time input variable
$\bm{r}$. To design the corresponding controller, the following premises are necessary.

\begin{itemize}
	\item {The focus of this paper is to design a controller to drive the end-effector of the manipulator moving to the desired position, and the problem of collision or trajectory planning is not the focus of this paper. The posture of the end-effector is not the point of our concern either.  }
	\item {In this paper, the camera is fixed at a specific position and posture, that is, the system is a typical eye-to-hand visual servoing system. Moreover, the depth camera can measure target’s position in camera coordination. The manipulator’s end is bound with an identified object, a red ball, for example, which makes image recognition easier (shown in Fig.1). We define the 3-dimensional position of the end-effector in camera coordination as system state which is written as:  
		\begin{equation}
			{\bm{x}}=  \left[
			\begin{array}{ccc}
				\bm{x}_{1} &
				\bm{x}_{2} &
				\bm{x}_{3}
			\end{array}
			\right]^{\mathrm{T}}.
	\end{equation}}
	\item {The manipulator is driven by the joint space velocity signal. The control signal is designed in the format of joint velocity $\dot{\bm{r}}  $.}
\end{itemize}

Definition 1\cite{23}: When it comes to a nonlinear system presented as:
\begin{equation}
	\dot{\bm{x}}= f(\bm{x},\bm{u}),f(0,0)=0.
\end{equation}
$\bm{x}$ denotes system state, and $\bm{u}$ denotes system input. The equilibrium point $\bm{x}=0$ is semi-global practical finite-time stable if for any initial state $\bm{x}(t_0)$,
there exist a positive scale $\mathrm{D}$ and corresponding time $T(\bm{x}(t_0),\mathrm{D})$ that satisfies $|\bm{x}|<\mathrm{D}$, when $t>T(\bm{x}(t_0),\mathrm{D})$.

Lemma 1\cite{24}: For variables $\psi$, $\xi$, and constant value $\iota$, $\upsilon$, and $\rho$, there is inequality:
\begin{equation}
	{\left| \psi  \right|^\upsilon }{\left| \xi  \right|^\rho } \le \frac{\upsilon }{{\upsilon  + \rho }}\iota {\left| \psi  \right|^{\upsilon  + \rho }} + \frac{\rho }{{\upsilon  + \rho }}{\iota ^{\frac{{ - \upsilon }}{\rho }}}{\left| \xi  \right|^{\upsilon  + \rho }}.
\end{equation}

Lemma 2\cite{25}: Let $ \textit{z}_{j} \in {\mathbb{R}} $, $ j=1,\dots,n $, $ 0<p  \le 1 $ , then inequality holds:
\begin{equation}
	{\left( {\sum\limits_{j = 1}^n {\left| {{z_j}} \right|} } \right)^p} \le \sum\limits_{j = 1}^n {{{\left| {{z_j}} \right|}^p}}  \le {n^{1 - p}}{\left( {\sum\limits_{j = 1}^n {\left| {{z_j}} \right|} } \right)^p}.
\end{equation}

Lemma 3\cite{23}: For the nonlinear system (2), suppose there is a smooth, positive function $V(\bm{x})$. When there are positive constant $k>0$, $0<\sigma<1$, and $\delta>0$ that satisfy the inequation:
\begin{equation}
	\dot V \le -kV_\sigma+\delta,
\end{equation}
then the system $\dot{\bm{x}}= f(\bm{x},\bm{u})$ is semi-global practical finite-time stable (SGPFS).

\section{Methodology}
\noindent The hand-eye relationship can be expressed in a local linear format between joint space velocity and characteristic velocity using a Jacobian matrix:
\begin{equation}
	\dot{\bm{x}}= \bm{J}\cdot\dot{\bm{r}}.
\end{equation}
$\dot {\bm{x}}$ denotes the velocity of the image feature; $\dot {\bm{r}}=[\dot {\bm{r}_1},...,\dot {\bm{r}_6}]^{\mathrm{T}}$ denotes the velocity of joint space. The Jacobian matrix $\bm{J}$ is time-varying due to the hand-eye relationship being highly coupled and nonlinear.

The key problem of Image-based Visual Servoing lies in the accuracy of estimation of the Jacobian matrix and the algorithm of control law, which can determine the convergency performance and the stability of the whole system\cite{16}. In this section, a radial-basis-function neural network is applied to approximate the current Jacobian matrix using the value of 6-dimensional joint space. In the offline stage, the RBFNN is trained based on the data collected via random movement of joint space. In the online stage, the RBFNN model will be updated, which will be proved to increase the robustness of the whole system. What’s more, a finite-time controller is applied to boost the convergence speed.

\subsection{Offline Training}
\noindent Considering the nonlinear trait of the hand-eye relationship, a RBFNN is applied for its global approximation capacity and strong nonlinear mapping ability\cite{26}. A serial of random joint space movement on the UR5 model is carried out to create initial training and test data sets.

The influence of six angel values on the end-effector position shows a decreasing trend due to the characteristic of the UR5 kinematic model, where six angel values correspond to different functions separately. For example, in the UR5 kinematic model, the axis corresponds to ${\bm{r}_1}$ serves as arm body rotation, impacting the end-effector position significantly. Whereas axis corresponds to ${\bm{r}_6}$ serve as end-effector rotation, showing no influence on the end-effector position. The decreasing trend of influence on the end-effector position from ${\bm{r}_1}$ to ${\bm{r}_6}$ manifests itself in the decreasing trend of the numeral size of six columns in the Jacobian matrix.

For the reasons above, the Jacobian matrix in formula (6) is split into six columns. The six columns separately indicate the influence of six angels on the end-effector position. It’s better for us to approximate the numerical value of six columns via six independent neural networks. Such method can accelerate the speed of cost function’s converging for applying different learning rates.
\begin{equation}
	\begin{aligned}
		\boldsymbol{J} &=\left[\begin{array}{ccc}
			\frac{\partial \boldsymbol{x}_{1}}{\partial \boldsymbol{r}_{1}} & \cdots & \frac{\partial \boldsymbol{x}_{1}}{\partial \boldsymbol{r}_{6}} \\
			\vdots & \ddots & \vdots \\
			\frac{\partial \boldsymbol{x}_{3}}{\partial \boldsymbol{r}_{1}} & \cdots & \frac{\partial \boldsymbol{x}_{3}}{\partial \boldsymbol{r}_{6}}
		\end{array}\right] \\
		&=\left[\begin{array}{llllll}
			\boldsymbol{J}_{1} & \boldsymbol{J}_{2} & \boldsymbol{J}_{3} & \boldsymbol{J}_{4} & \boldsymbol{J}_{5} & \boldsymbol{J}_{6}
		\end{array}\right]
	\end{aligned}
\end{equation}

The specific value of the ${{t}^{\rm{th}}}$ hidden neuron of the ${i^{\rm{th}}}$ RBFNN as well as the estimated ${\hat {\bm{J}}_{{i}}}$ are as follows.
\begin{equation}
	{\bm{\theta} _{{{it}}}} = {{\rm{e}}^{{\rm{ - }}\left(  \frac{{||\bm{r} - {\bm{u}_{{{it}}}}||}}{{{\bm{\delta} _{{{it}}}}}}\right)  {^2}}}  = {{\rm{e}}^{{\rm{ - }}\left[ \sqrt {\sum\limits_{{{j}} = 1}^{\rm{l}} {{{(\frac{{{\bm{r}_{{j}}} - {\bm{u}_{{{itj}}}}}}{{{\bm{\delta} _{{{it}}}}}})}^2}} }\right]  {^2}}}
\end{equation}
\begin{equation}
	{\hat {\bm{J}}_{{i}}} = {\bm{W}_{{i}}}{\bm{\theta} _{{i}}}(\bm{r}),{{i}} = 1,...,6
\end{equation}
Where the parameters ${\bm{u}_{{{it}}}}$ and ${\bm{\delta} _{{{it}}}}$ are trained in offline stage, and ${\bm{W}_{{i}}}$ is the matrix of weights of the $i^{\mathrm{th}}$ RBFNN. By introducing the divided form of the Jacobian matrix into $\dot{\bm{x}}= \bm{J}\cdot\dot{\bm{r}}$, the velocity of characteristic is expressed as:
\begin{equation}
	\begin{array}{ccccc}
		\dot{ \bm{x} }= \!\! & \left[ {\begin{array}{*{20}{c}}
				{{\bm{J}_1}}& \cdots &{{\bm{J}_6}}
		\end{array}} \right]\left[ {\begin{array}{*{20}{c}}
				{{{\dot{ \bm{r}}}_1}}\\
				\vdots \\
				{{{\dot { \bm{r}}}_6}}
		\end{array}} \right] \!\!
		= \!\! & \sum\limits_{{{i}} = 1}^6 {{\bm{W}_{{i}}} \cdot {\bm{\theta} _{{i}}} \cdot {{\dot{\bm{r}} }_{{i}}}} .
	\end{array}
\end{equation}
Here we define the error of estimation of characteristic speed as:
\begin{equation}
	\bm{e} = \dot{\bm{x}} - \hat{\bm{J}}\dot{\bm{r}} = \sum\limits_{{{i}} = 1}^6 {{\bm{W}_{{i}}}{\bm{\theta} _{{i}}}{{\dot {\bm{r}}}_{{i}}}}  - \sum\limits_{{{i}} = 1}^6 {{{\hat {\bm{W}}}_{{i}}}{\bm{\theta} _{{i}}}{{\dot {\bm{r}}}_{{i}}}}  = \sum\limits_{{{i}} = 1}^6 {\Delta {{\bm{W}}_{{i}}}{\bm{\theta} _{{i}}}{{\dot{\bm{ r}}}_{{i}}}} .
\end{equation}
The offline model only constructs an estimation algorithm of the Jacobian matrix, thus forming the local linear model of the hand-eye relationship. The controller is designed in the following section with the estimated Jacobian matrix.

\subsection{Finite-time Controller Design and Stability Proof}
\noindent In this section, the control law and update algorithm designed with sliding mode control theory will be discussed, and the corresponding stability proof will be provided. Few previous visual servoing works related to RBFNN have achieved finite-time control results. A more efficient controller with a finite-time control effect will be provided later.
\subsubsection{Controller}
Sliding mode control is essentially a kind of special nonlinear control whose control procedure can be divided into two stages\cite{27}. The first stage is movement from the initial state to the sliding surface, usually called the arrival stage. The second stage is movement on the sliding surface leading to the equilibrium point, called convergency stage\cite{28}. Sliding mode control is a control algorithm with proven merits of strong robustness and excellent anti-interference performance\cite{29,30}. Sliding mode control is suitable for the demanding visual servoing task sensitive to environment.

The primitive linear and integral sliding surface is usually designed as:
\begin{equation}
	\bm{s}{\text{ = }}{{\text{c}}_1}\Delta \bm{x} + {{\text{c}}_2}\Delta \dot {\bm{x}}
\end{equation}
\begin{equation}
	\bm{s}{\text{ = }}{{\text{c}}_1}\Delta \bm{x} + {{\text{c}}_2}\int {\Delta \bm{x}{\text{d}t}} 
\end{equation}
Where the positive scale $c_1$ and $c_2$ serve as sliding surface parameters. The term $\Delta \bm{x}=\bm{x}-\bm{x}_{\rm{d}}$ indicates error in characteristics, and $\bm{x}_{\rm{d}}$ indicates the desired characteristics. The linear term $c_1\Delta \bm{x}$ and $c_2\Delta \dot {\bm{x}}$ is set by the system state, whereas the integral term ${{\text{c}}_2}\int {\Delta \bm{x}{\text{dt}}}$ can be set autonomously. Once the integral term in (13) is set as
$\int {\Delta \bm{x}{\text{d}t}}=-\frac{c_2}{c_2}\Delta \bm{x}$, the system state is on the sliding surface at the very beginning, thus accelerating the convergency procedure. When system state is on sliding surface, there is $\bm{s}=0$ and $\dot {\bm{s}}=0$. When it comes to the convergency stage, introduce $\bm{s}=0$ and $\dot {\bm{s}}=0$ respectively into (12) and (13), then the rate of convergence can be evaluated via $\bm{s}=c_1\Delta \bm{x}+c_2\Delta \dot {\bm{x}}=0$ and $\dot {\bm{s}}=c_1\Delta \dot {\bm{x}}+c_2\Delta \bm{x}=0$ respectively. It is obvious that both of them converge exponentially. Traditional sliding mode control applies for a linear or integral sliding surface whose convergency speed is considerably slow when the state is close to the equilibrium point, leading to an imponderable convergency time. 

Fast Terminal Sliding Mode (FTSM) control was proposed to address the shortcoming of imponderable converging time, where both the linear proportion and terminal attractor are applied. The design ensures the rate of convergence when it is near or far away from the equilibrium point\cite{31}. In addition, the convergency time of the system state moving on the sliding surface can be calculated. A corresponding control law and update algorithm based on FTSM will be discussed.

A typical fast terminal sliding surface can be designed as:
\begin{equation}
	\bm{s}{\text{ = }}{\alpha _1}\vartriangle \dot {\bm{x}} + {\alpha _2}\vartriangle \bm{x} + {\alpha _3}{ \operatorname{abs}({\vartriangle \bm{x}}) ^{\bm{\gamma}} }\circ\operatorname{sgn} \left( {\vartriangle \bm{x}} \right)
\end{equation}
Where $\alpha_1$, $\alpha_2$, and $\alpha_3$ are positive scale serving as sliding surface parameters. The $\operatorname{abs}({\vartriangle \bm{x}})$ function represents taking the absolute value of each element of the vector ${\vartriangle \bm{x}}$. and the elements in vector ${\bm{\gamma}}$ act as exponent. The derivative of the sliding surface can be obtained as:
\begin{equation}
	\dot {\bm{s}} = {\alpha _1}\vartriangle \ddot {\bm{x}} + {\alpha _2}\vartriangle \dot {\bm{x}} + {\alpha _3}{\bm{\gamma}} { \operatorname{abs}({\vartriangle \bm{x}}) ^{{\bm{\gamma}}  - 1}}\circ\vartriangle \dot {\bm{x}}.
\end{equation}

The control law applied in joint space is designed as:
\begin{equation}
	\dot {\bm{r}} = \frac{1}{{{\alpha _2}}}{(\hat {\bm{J}})^ + }\left[ { - {k_1}\bm{s} - {k_2}{{ \operatorname{abs}(\bm{s}) }^{2\sigma  - 1}} \circ \operatorname{sgn} \left( \bm{s} \right) - {k_4}\dot {\bm{s}} + {\alpha _2}\vartriangle \dot {\bm{x}}} \right]
\end{equation}
The term $(\hat {\bm{J}})^ +$ applied here indicates the Moore-Penrose pseudo-inverse of the estimated Jacobian matrix. The weights matrix of RBFNN is updated with data collected via calculation and observation. And the update law for the ${j^{\rm{th}}}$ row of the ${i^{\rm{th}}}$ weights matrix is designed as:
\begin{equation}
	{\dot {\hat {\bm{W}}}}^{\mathrm{T}}_{{{ij}}} = {\dot {\bm{r}}_{{i}}}{\bm{\theta} _{{i}}}({{\text{n}}_1}{{\bm{e}}_{{j}}} - {\alpha _2}{{\bm{s}}_j}) - {k_3}{\hat {\bm{W}}^{\mathrm{T}}}_{{{ij}}}
\end{equation}
$\text{n}_1$ is a positive number and the term $\bm{e}_{{j}}$ corresponds to the elements of the error vector of characteristic speed estimation. It is transparent from (16) and (17) that the information of the sliding surface is included in the RBFNN weight matrix update law as well as the joint space control law.
\subsubsection{Stability Proof}
The controller in this paper (represented from (14) to (17)) are designed to make a corresponding Lyapunov function  $ {{V}}$ go down numerically, which helps the stability discuss. Basically, the form of Lyapunov function in relating research can be divided into two categories. One kind of Lyapunov function only integrates current feedback error, holding the assumption that corresponding estimation for Jacobian matrix using theory like Kalman Filter or BFGS update algorithm is reliable, which may make its stability proof less than perfect. The other kind of Lyapunov function contains information about current feedback error and the error of the estimated Jacobian matrix, which means that the error information of both system output and estimated Jacobian matrix in closed-loop system are fully taken into consideration when designing controller and update algorithm.
The control law and the update law are supposed to ensure the system’s stability. The semi-global practical finite-time stability of the whole system is analyzed as follows. By multiplying both sides of (16) by $  \bm{\hat J}$. It can be obtained that:
\begin{small}
	\begin{equation}
		\hat {\bm{J}}\dot {\bm{r}} = \frac{1}{{{\alpha _2}}}\hat {\bm{J}}{(\hat {\bm{J}})^ + }\left[ { - {k_1}{\bm{s}} - {k_2}{{\operatorname{abs}(\bm{s})}^{2\sigma  - 1}} \circ \operatorname{sgn} \left( \bm{s} \right) - {k_4}\dot {\bm{s}} + {\alpha _2}\Delta \dot {\bm{x}}} \right]
	\end{equation}
\end{small}
By introducing $\dot{\bm{x}}= \bm{J}\cdot\dot{\bm{r}} $ and $\bm{e}= \dot{\bm{x}}- \bm{\hat J} \bm{\dot r }$ into (11), the relationship between $\bm{e}$ and $\dot{\bm{x}}$ can be represented as:
\begin{equation}
	\hat{\bm{J}}\dot {\bm{r}} = \hat {\bm{J}}\dot {\bm{r}} - \bm{J}\dot {\bm{r}} + \bm{J}\dot {\bm{r}} =  - \bm{e} + \dot {\bm{x}}.
\end{equation}
Introducing $\left( 18\right) $ into (19), there is
\begin{small}
	\begin{equation}
		\bm{e} = \dot {\bm{x}} - \frac{1}{{{\alpha _2}}}\hat {\bm{J}}{(\hat {\bm{J}})^ + }\left[ { - {k_1}\bm{s} - {k_2}{{\operatorname{abs}(\bm{s})}^{2\sigma  - 1}} \circ \operatorname{sgn} \left( \bm{s} \right) - {k_4}\dot {\bm{s}} + {\alpha _2}\Delta \dot {\bm{x}}} \right]
	\end{equation}
\end{small}
To verify the stability of the system, a specific Lyapunov function is designed as:
\begin{equation}
	{{V}} = \frac{1}{2}{k_4}{\bm{s}^{\mathrm{T}}}\bm{s} + \frac{1}{2}\sum\limits_{{{i}} = 1}^{{n}} {\sum\limits_{{{j}} = 1}^{{l}} {\Delta {\bm{W}_{{{ij}}}}\Delta {\bm{W}}_{{{ij}}}^{\mathrm{T}}} } 
\end{equation}

In this part, the Lyapunov function consists of information of both system error and error of estimated Jacobian matrix, and the latter is presented in the form of error of weight matrix. Such method has been applied in many cases, and corresponding stability has been rigorously proved\cite{17}. However, relevant stability proof usually merely includes a numerical decrease, which usually leads to a infinite convergence time in principle. Inspired by the application of SGPFS theory in visual servoing task in \cite{1}, where the Lyapunov function satisfies the differential inequality in (5), meaning that the corresponding Lyapunov function is decreasing so that it has an upper boundary. By adjusting the parameter, we can make sure that the function asymptotically converges to a compact set around zero in a corresponding finite time.
Considering (14), (16), and (17), the derivation of Lyapunov function is calculated as:
\begin{equation}
	\begin{aligned}
		{\dot V} =  & {k_4}{\bm{s}^{\mathrm{T}}}\dot {\bm{s}} - \sum\limits_{{{i}} = 1}^{{n}} {\sum\limits_{{{j}} = 1}^{{l}} {\Delta {\bm{W}_{{{ij}}}}{\dot {\hat {\bm{W}}}}_{{{ij}}}^{\mathrm{T}}} }  \\ 
		=  & {\bm{s}^{\mathrm{T}} }\left( {{k_4}\dot {\bm{s}} - {\alpha _2}\Delta \dot {\bm{x}} + {\alpha _2}\bm{e} + {\alpha _2}\hat {\bm{J}}\dot{\bm{r}}} \right) \\
		& - \sum\limits_{{{i}} = 1}^{{n}} {\sum\limits_{{{j}} = 1}^{{l}} {\Delta {\bm{W}_{{{ij}}}}\left[ {{{\dot {\bm{r}}}_{{i}}}{\bm{\theta} _{{i}}}({{{n}}_1}{\bm{e}_{{j}}} + {\alpha _2}{\bm{s}_j}) - {k_3}{{\hat {\bm{W}}}^{\mathrm{T}}}_{{{ij}}}} \right]} }  \\ 
		=  & {\bm{s}^{\mathrm{T}} }\left( { - {k_1}\bm{s} - {k_2}{{\operatorname{abs}(\bm{s})}^{2\sigma  - 1}} \circ \operatorname{sgn} \left( \bm{s} \right) + {\alpha _2}\bm{e}} \right) \\
		& - \sum\limits_{{{i}} = 1}^{{n}} {\sum\limits_{{{j}} = 1}^{{l}} {\Delta {\bm{W}_{{{ij}}}}\left[ {{{\dot {\bm{r}}}_{{i}}}{\bm{\theta} _{{i}}}({{{n}}_1}{\bm{e}_{{j}}} + {\alpha _2}{\bm{s}_j}) - {k_3}{{\hat {\bm{W}}}^{\mathrm{T}}}_{{{ij}}}} \right]} }  \\ 
		=  &  - {k_1}{\bm{s}^{\mathrm{T}} }\bm{s} - {k_2}{\left( {{\bm{s}^\mathrm{T} }\bm{s}} \right)^\sigma } - {{{n}}_1}{\bm{e}^\mathrm{T}}\bm{e}  \\
		& + {k_3}\sum\limits_{{{i}} = 1}^{{n}} {\sum\limits_{{{j}} = 1}^{{l}} {\Delta {\bm{W}_{{{ij}}}}{{\hat {\bm{W}}}^\mathrm{T}}_{{{ij}}}} }  \\ 
	\end{aligned} 
\end{equation}
After arranging the formula in (22), we can get
\begin{equation}
	{\dot V} \le    - \frac{{{k_2}}}{{k_4^\sigma }}{\left( {{k_4}{\bm{s}^\mathrm{T} }\bm{s}} \right)^\sigma } + {k_3}\sum\limits_{{{i}} = 1}^{{n}} {\sum\limits_{{{j}} = 1}^{{l}} {\Delta {\bm{W}_{{{ij}}}}{{\hat {\bm{W}}}^\mathrm{T}}_{{{ij}}}} } 
\end{equation}
Meanwhile, from the definition of ${{\hat W}^{T}}_{{{ij}}}$, it can be obtained that:
\begin{equation}
	\Delta {\bm{W}_{{{ij}}}}{\hat {\bm{W}}^\mathrm{T}}_{{{ij}}}  \le  - \frac{1}{2}\Delta {\bm{W}_{{{ij}}}}\Delta {\bm{W}_{{{ij}}}}^\mathrm{T} + \frac{1}{2}{\bm{W}_{{{ij}}}}{\bm{W}_{{{ij}}}}^\mathrm{T}
\end{equation}
Substituting (24) into (23), one has:
\begin{equation}
	\begin{aligned}
		{\dot V} \le & - \frac{{{k_2}}}{{k_4^\sigma }}{\left( {{k_4}{\bm{s}^\mathrm{T} }\bm{s}} \right)^\sigma } - \frac{{{k_3}}}{2}\sum\limits_{{{i}} = 1}^{{n}} {\sum\limits_{{{j}} = 1}^{{l}} {\Delta {\bm{W}_{{{ij}}}}\Delta {\bm{W}_{{{ij}}}}^\mathrm{T}} } \\
		&+ \frac{{{k_3}}}{2}\sum\limits_{{{i}} = 1}^{{n}} {\sum\limits_{{{j}} = 1}^{{l}} {{\bm{W}_{{{ij}}}}{\bm{W}_{{{ij}}}}^\mathrm{T}} } 
	\end{aligned}
\end{equation}
Applying Lemma 1, let $\psi=1$, $\xi={\Delta {\bm{W}_{{{ij}}}}\Delta {\bm{W}_{{{ij}}}}^\mathrm{T}}$, $\upsilon  = 1 - \sigma$, $\rho  = \sigma$, $\iota  = {\sigma ^{\frac{\sigma }{{1 - \sigma }}}}$, one has:
\begin{equation}
	{\left( {\Delta {\bm{W}_{{{ij}}}}\Delta {\bm{W}_{{{ij}}}}^\mathrm{T}} \right)^\sigma } \le \left( {1 - \sigma } \right)\iota  + \Delta {\bm{W}_{{{ij}}}}\Delta {\bm{W}_{{{ij}}}}^\mathrm{T}
\end{equation}
Combining (25) and (26), there is:
\begin{equation}
	\begin{aligned}
		{{\dot V}} \le &    \frac{{{k_3}}}{2}\sum\limits_{{{i}} = 1}^{{n}} {\sum\limits_{{{j}} = 1}^{{l}} {\left[ {\left( {1 - \sigma } \right)\iota  - {{\left( {\Delta {\bm{W}_{{{ij}}}}\Delta {\bm{W}_{{{ij}}}}^\mathrm{T}} \right)}^\sigma }} \right]} }  \\
		&+ \frac{{{k_3}}}{2}\sum\limits_{{{i}} = 1}^{{n}} {\sum\limits_{{{j}} = 1}^{{l}} {{\bm{W}_{{{ij}}}}{\bm{W}_{{{ij}}}}^\mathrm{T}} } - \frac{{{k_2}}}{{k_4^\sigma }}{\left( {{k_4}{\bm{s}^\mathrm{T} }\bm{s}} \right)^\sigma } \\ 
		\le & - \frac{{{k_2}}}{{k_4^\sigma }}{\left( {{k_4}{\bm{s}^\mathrm{T} }\bm{s}} \right)^\sigma } - \frac{{{k_3}}}{2}\sum\limits_{{{i}} = 1}^{{n}} {\sum\limits_{{{j}} = 1}^{{l}} {{{\left( {\Delta {\bm{W}_{{{ij}}}}\Delta {\bm{W}_{{{ij}}}}^\mathrm{T}} \right)}^\sigma }} }  \\
		&+ \frac{{{k_3}}}{2}\sum\limits_{{{i}} = 1}^{{n}} {\sum\limits_{{{j}} = 1}^{{l}} {\left( {1 - \sigma } \right)\iota } }  + \frac{{{k_3}}}{2}\sum\limits_{{{i}} = 1}^{{n}} {\sum\limits_{{{j}} = 1}^{{l}} {{\bm{W}_{{{ij}}}}{\bm{W}_{{{ij}}}}^\mathrm{T}} }  \\ 
		\le & - \frac{{{k_2}}}{{k_4^\sigma }}{\left( {{k_4}{\bm{s}^\mathrm{T} }\bm{s}} \right)^\sigma } - k\sum\limits_{{{i}} = 1}^{{n}} {\sum\limits_{{{j}} = 1}^{{l}} {{{\left( {\Delta {\bm{W}_{{{ij}}}}\Delta {\bm{W}_{{{ij}}}}^\mathrm{T}} \right)}^\sigma }} }  \\
		&+ \frac{{{k_3}}}{2}\sum\limits_{{{i}} = 1}^{{n}} {\sum\limits_{{{j}} = 1}^{{l}} {\left( {1 - \sigma } \right)\iota } }  + \frac{{{k_3}}}{2}\sum\limits_{{{i}} = 1}^{{n}} {\sum\limits_{{{j}} = 1}^{{l}} {{\bm{W}_{{{ij}}}}{\bm{W}_{{{ij}}}}^\mathrm{T}} }  \\ 
	\end{aligned} 
\end{equation}
Where $k = \min \left\{ { - \frac{{{k_2}}}{{k_4^\sigma }},\frac{{{k_3}}}{2}} \right\}$. Then applying Lemma 2, one has:
\begin{equation}
	\begin{aligned}
		{{\dot V}} \le  &  - k{\left( {{\bm{s}^\mathrm{T} }\bm{s} + \sum\limits_{{{i}} = 1}^{{n}} {\sum\limits_{{{j}} = 1}^{{l}} {\Delta {\bm{W}_{{{ij}}}}\Delta {\bm{W}_{{{ij}}}}^\mathrm{T}} } } \right)^\sigma } \\
		&+ \frac{{{k_3}}}{2}\sum\limits_{{{i}} = 1}^{{n}} {\sum\limits_{{{j}} = 1}^{{l}} {\left( {1 - \sigma } \right)\iota } }  + \frac{{{k_3}}}{2}\sum\limits_{{{i}} = 1}^{{n}} {\sum\limits_{{{j}} = 1}^{{l}} {{\bm{W}_{{{ij}}}}{\bm{W}_{{{ij}}}}^\mathrm{T}} }  \\ 
		\le  &  - k{{{V}}^\sigma } + \delta  \\ 
	\end{aligned} 
\end{equation}
Where 
$\delta  = \frac{{{k_3}}}{2}\sum\limits_{{{i}} = 1}^{{n}} {\sum\limits_{{{j}} = 1}^{{l}} {\left( {1 - \sigma } \right)\iota } }  + \frac{{{k_3}}}{2}\sum\limits_{{{i}} = 1}^{{n}} {\sum\limits_{{{j}} = 1}^{{l}} {{\bm{W}_{{{ij}}}}{\bm{W}_{{{ij}}}}^\mathrm{T}} } $.  
The differential equation in (28) indicates that the signals in the closed-loop system are semi-global practical finite-time stable. Then, let 
${T^ * } = \frac{1}{{\left( {1 - \sigma } \right)\varphi k}}\left[ {{V^{1 - \sigma }}\left( {\bm{x}\left( 0 \right)} \right) - {{\left( {\frac{\delta }{{\left( {1 - \varphi } \right)k}}} \right)}^{{{\left( {1{{ - }}\sigma } \right)} \mathord{\left/{\vphantom {{\left( {1{{ - }}\sigma } \right)} \sigma }} \right.\kern-\nulldelimiterspace} \sigma }}}} \right]$.
From Lemma 3, for $\forall {{t}} \ge {{{T}}^ * }$, there is 
$V\left( \bm{x} \right) \le {\left[ {\frac{\delta }{{\left( {1 - \varphi } \right)k}}} \right]^{\frac{1}{\sigma }}}$.
It means that the system states can approach the sliding surface with any specific error in finite time by selecting appropriate parameters.

The analysis of finite-time convergence property is divided into two parts. The first part is the process of the system state convergence to the sliding surface in finite time. The second part is the process of the system state convergence to the equilibrium point on sliding surface in finite time. In the previous paragraph, the Lyapunov function which consists of ${\bm{s}^{\mathrm{T}}}\bm{s}$ and ${\Delta {\bm{W}_{{{ij}}}}\Delta {\bm{W}}_{{{ij}}}^{\mathrm{T}}}$ can reach to a desired neighborhood of zero in finite time, which means that system state can move to the desired neighborhood of sliding surface in finite time, considering the form of Lyapunov function.
\subsubsection{Adaptive Factor}
The proposed controller is numercially proved to acquire the effect of finite-time control. However, there is still room for improvement. In order to introduce following content, a simple numerical simulation experiment is carried out to verify the control algorithm's reliability. And the experiment result is shown in Fig.2.

\begin{figure}[htbp]
	\centering
	\includegraphics[width=6cm]{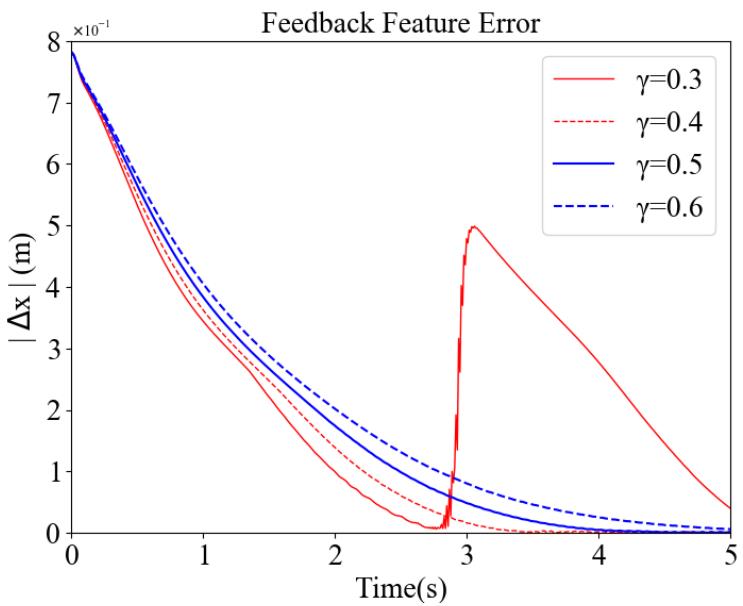}
	\caption{Convergency plot corresponding to different exponents. ${\bm{\gamma}} = 0.3$ means that ${{\bm{\gamma}}_j}=0.3,j=1,2,3.$}
\end{figure}

In terms of system stability, it is obvious that the system state is unstable when the exponent is too small. The convergency attractor accelerates the rate of convergency. However, it introduces a singularity situation. There are large fluctuations in the system state for the existence of a negative exponent in (15). It is noted that $0< {{\bm{\gamma}}_j} <1$. When the system error is too close to 0, the controller produces a considerably large value, making the system unstable. In this experiment, when the exponent ${{\bm{\gamma}}_j} \ge 0.4$, fluctuation gets alleviated. 

In terms of the rate of convergency, it is shown in Fig.2 that when the error distance $|\Delta \bm{x}_j|$ is less than 1, the smaller exponent can speed up the system’s converging. However, a small exponent may cause unexpected mutations in system state, so it is hard to find a balance between stability and convergency performance. Therefore, a desired ${\bm{\gamma}}_j$ is supposed to be a variable that varies with the corresponding error $|\Delta \bm{x}_j|$. When the error is large, a relatively small ${\bm{\gamma}}_j$ is desired to ensure a faster rate of convergence. When the error is small, a relatively large ${\bm{\gamma}}_j$ is desired to pursue the stability of system. In this way, both the convergence performance and robustness can be guaranteed. 

The variant exponent can be set as:
\begin{equation}
	{\bm{\gamma}_j} \left( {\Delta \bm{x}_j} \right) = {\lambda _1} - {\lambda _2} \cdot \tanh \left[ {{\lambda _3}\left( {\Delta {{\bm{x}_j}^2} - \Delta } \right)} \right]
\end{equation}
Where ${{{\lambda}}}_1$, ${{{\lambda}}}_2$, and ${{{\lambda}}}_3$ are all positive scale. $\Delta$ is a small positive constant. Apparently, the ${\bm{\gamma}_j} \left( \Delta \bm{x}_j \right) $ in (29) is a typical switch function, and the switch motion takes place when $|\Delta \bm{x}_j| \approx \sqrt{\Delta }$. The switching rate depends on ${\lambda}_3$. The stable state of this function relies on ${\lambda}_1$ and ${\lambda}_2$. In addition, when ${\lambda}_3 \gg 1$, one has
\begin{equation}
	{\bm{\gamma}_j} \left( {\Delta \bm{x}_j} \right)\left\{ {\begin{array}{*{20}{c}}
			{ \to {{\lambda} _1} - {{\lambda} _2}}&{\left| {\Delta \bm{x}_j} \right| > \sqrt \Delta  } \\ 
			{ \to {{\lambda} _1} - {{\lambda} _2} \cdot \tanh \left[ {{{\lambda} _3}\left( { - \Delta } \right)} \right]}&{\left| {\Delta \bm{x}_j} \right| < \sqrt \Delta  } 
	\end{array}} \right.
\end{equation}
Then, we get the sliding surface and its derivative:
\begin{equation}
	\bm{s}{{ = }}{\alpha _1}\Delta \dot {\bm{x}} + {\alpha _2}\Delta \bm{x} + {\alpha _3}{\operatorname{abs}( {\Delta \bm{x}} )^{{\bm{\gamma}} \left( {\Delta \bm{x}} \right)}}\circ \operatorname{sgn} \left( {\Delta \bm{x}} \right)
\end{equation}
\begin{equation}
	\begin{aligned}
		&\dot {\bm{s}} = {\alpha _1}\Delta \ddot {\bm{x}} + {\alpha _2}\Delta \dot {\bm{x}} + {\alpha _3}[  {{\operatorname{abs}( {\Delta \bm{x}} )}^{{\bm{\gamma}} \left( {\Delta \bm{x}} \right){{ - }}1}}\circ{\bm{\gamma}} \left( {\Delta \bm{x}} \right)\\   
		& +{{\operatorname{abs}( {\Delta \bm{x}} )}^{\bm{\gamma} \left( {\Delta \bm{x}} \right)}} \circ \dot {\bm{\gamma}} \left( {\Delta \bm{x}} \right)\circ\ln (\operatorname{abs}( {\Delta \bm{x}} ))]\circ \operatorname{sgn} \left( {\Delta \bm{x}} \right) 
	\end{aligned}
\end{equation}
The new update law for the weights matrix is the same as formula in (17). And the control law is identical in form to the original one in (16), except for the variant exponent in $\dot{\bm{s}}$ and $\bm{s}$. And the Lyapunov function still satisfies the following differential equation:
\begin{equation}
	{{\dot V}} \le   - k{{{V}}^\sigma } + \delta 
\end{equation}
Where \begin{small}$\delta  = \frac{{{k_3}}}{2}\sum\limits_{{{i}} = 1}^{{n}} {\sum\limits_{{{j}} = 1}^{{l}} {\left( {1 - \sigma } \right)\iota } }  + \frac{{{k_3}}}{2}\sum\limits_{{{i}} = 1}^{{n}} {\sum\limits_{{{j}} = 1}^{{l}} {{\bm{W}_{{{ij}}}}{\bm{W}_{{{ij}}}}^\mathrm{T}} }$\end{small}, and $0<\sigma<1$.

Remark: To accelerate system convergence, it is recommended to select a ${\bm{\gamma}}_j>1$ when $|\Delta \bm{x}_j|>1$ and select a ${\bm{\gamma}}_j<1$ when $|\Delta \bm{x}_j | <1$. However, long-distance visual servoing on manipulator may involve collisions and planning problem, which are not the focus of this article. Therefore, the theoretical derivation and experiment of this paper are designed under the premise that initial distance \begin{small}$|\Delta \bm{x}_j\left( 0\right) |<1$\end{small}.

To illustrate the role variant exponent plays in fetching a balance between convergency speed and system stability. A simple numerical experiment is carried out to verify the control algorithm's reliability and find suitable sliding surface parameters and controller parameters. The result is shown in Fig.3.
\begin{figure}[htbp]
	\centering
	\includegraphics[width=8.8cm]{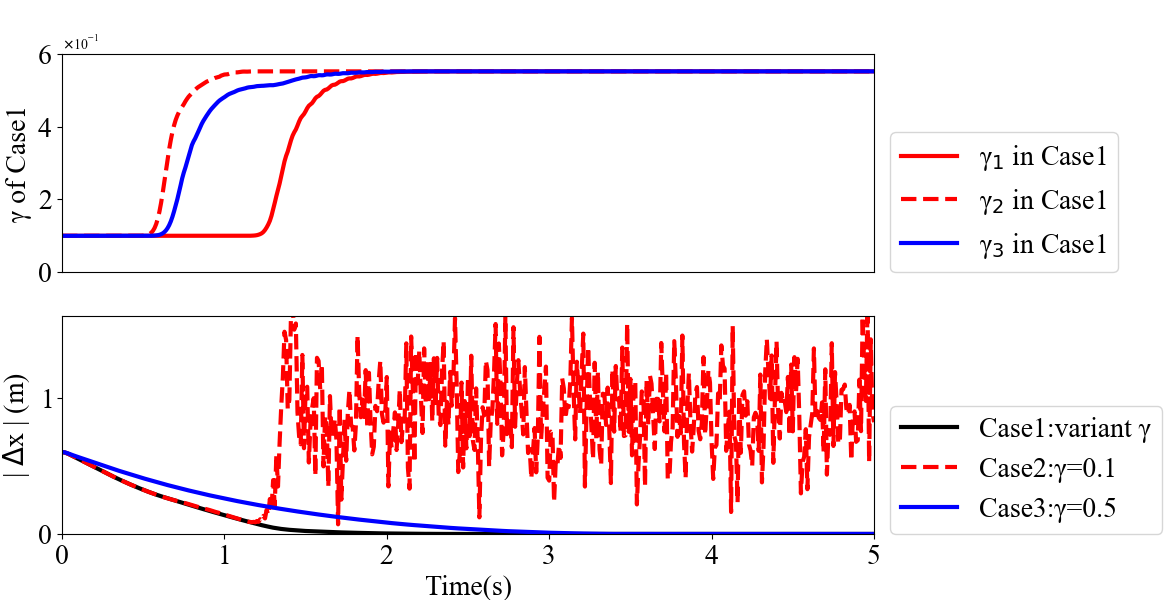}
	\caption{Profiles of the criteria error corresponding to different exponents and profiles of the variant exponents in $\mathrm{Case}1$. ${\bm{\gamma}} = 0.1$ means that ${{\bm{\gamma}}_j}=0.1,j=1,2,3.$}
\end{figure}
The value of $\bm{\gamma}_j$ switches between ${\lambda}_1-{\lambda}_2 \cdot \mathrm{tanh}\left[ {\lambda}_3\left( \Delta {\bm{x}_j}^2 \left( 0\right) -\Delta \right) \right] $ 
and ${\lambda}_1-{\lambda}_2 \cdot \mathrm{tanh}\left[ {\lambda}_3\left(  -\Delta \right) \right] $  
during control procedure. When system error is relatively large, ${\bm{\gamma}}_j \approx {\lambda}_1-{\lambda}_2 = 0.1 , j=1,2,3$
And the system’s convergency performance is similar to that in $\mathrm{Case}_2$ whose exponent is set at 0.1. When system error is relatively small, 
${\bm{\gamma}}_j \approx {\lambda}_1-{\lambda}_2 \cdot \mathrm{tanh}\left[ {\lambda}_3\left(  -\Delta \right) \right] \approx 0.5 $. 
The relatively large $ {\bm{\gamma}}_j$ smooths convergency plot, and system in possess relatively ideal stability similar to that in $\mathrm{Case}_3$ whose exponent is set larger at 0.5.

The block diagram of the proposed framework is given in Fig.4.

\begin{figure}[htbp]
	\centering
	\includegraphics[width=12cm]{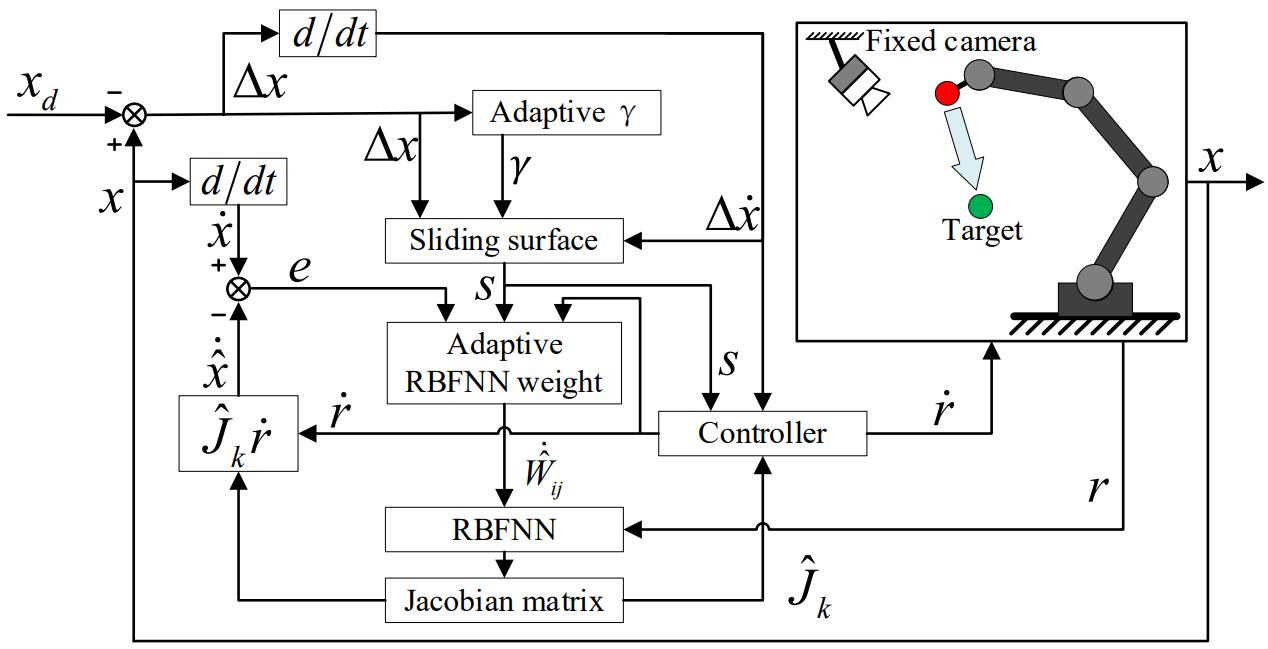}
	\caption{The block diagram of the control algorithm proposed in this paper.}
\end{figure}

\subsection{Convergence Speed Analysis}
\noindent The control procedure from the initial state to the equilibrium point can be divided into two stages. One is approach stage, in which the system state is moving towards the sliding surface. The other is convergency stage, in which the system state is moving on the switching surface, meaning that there is $\bm{s}_j=0$ as well as $\dot{ \bm{s}}_j=0$ in this stage.

It is exploited from (29) that sliding surface parameters ${\bm{\gamma}}_j$ are approximately constant before and after their value mutations. For the Lyapunov function satisfying the differential inequation
$	{{\dot V}} \le   - k{{{V}}^\sigma } + \delta $, from Lemma 4 and the definition of $\bm{s}$, when 
$t \ge t_s, t_s = \frac{1}{{\left( {1 - \sigma } \right)\varphi k}}\left[ {{V^{1 - \sigma }}\left( {\bm{x}\left( 0 \right)} \right) - {{\left( {\frac{\delta }{{\left( {1 - \varphi } \right)k}}} \right)}^{{{\left( {1{{ - }}\sigma } \right)} \mathord{\left/{\vphantom {{\left( {1{{ - }}\sigma } \right)} \sigma }} \right.\kern-\nulldelimiterspace} \sigma }}}} \right]$, there is 
$\left| {{\bm{s}_j}} \right| \le {\left[ {\frac{\delta }{{\left( {1 - \varphi } \right)k}}} \right]^{\frac{1}{{2\sigma }}}}$.It means that the system states can approach the sliding surface with any specific error in finite time by adjusting $\delta$, $\varphi$, and $k$. By selecting appropriate parameters, the system state can be limited to a considerably small area near the sliding surface.  When the system state is on the sliding surface, the convergency time can be evaluated with sliding surface information. The sliding surface is designed as:
\begin{equation}
	{\bm{s}_j}{{ = }}{\alpha _1}\Delta {\dot {\bm{x}}_j} + {\alpha _2}\Delta {\bm{x}_j} + {\alpha _3}{\left| {\Delta {\bm{x}_j}} \right|^{\bm{\gamma} \left( {\Delta {\bm{x}_j}} \right)}}\operatorname{sgn} \left( {\Delta {\bm{x}_j}} \right)
\end{equation}
The exponent $\bm{\gamma}_{j}$ varies when $\Delta \bm{x}_{j}$ changes. Therefore, the convergency time should be discussed case by case. Because the traits that ${\lambda}_3 \gg 1$, there is 
\begin{equation}
	{\bm{\gamma}_j} \left( {\Delta \bm{x}_j} \right)\left\{ {\begin{array}{*{20}{c}}
			{ \to {{\lambda} _1} - {{\lambda} _2}}&{\left| {\Delta \bm{x}_j} \right| > \sqrt \Delta  } \\ 
			{ \to {{\lambda} _1} - {{\lambda} _2} \cdot \tanh \left[ {{{\lambda} _3}\left( { - \Delta } \right)} \right]}&{\left| {\Delta \bm{x}_j} \right| < \sqrt \Delta  } 
	\end{array}} \right.
\end{equation}
Let ${t_{{\text{r}}j1}}$ correspond to the time it takes for the variable ${\bm{x}_j}$ to converge from ${\bm{x}_j(t_s)}$ to range $[-\sqrt{\Delta},\sqrt{\Delta}]$.
Let ${t_{{\text{r}}j2}}$ correspond to the time it takes for the variable ${\bm{x}_j}$ to converge from ${\bm{x}_j(t_s+t_{{\text{r}}j1})}$ to 0.
The discussion can be divided into two parts generally.
\begin{enumerate}
	\item {If $|\Delta \bm{x}_j(t_s)| > \sqrt{\Delta}$, there is
		${\bm{s}_j}{{ = }}{\alpha _1}\Delta {\dot {\bm{x}}_j} + {\alpha _2}\Delta {\bm{x}_j} + {\alpha _3}{\left| {\Delta {\bm{x}_j}} \right|^{{{\lambda} _1} - {{\lambda} _2}}}\operatorname{sgn} \left( {\Delta {\bm{x}_j}} \right)$.
		And the convergency time can be exploited that:
		\begin{equation}
			{t_{{\text{r}}j1}} \le  \frac{{{\alpha _1 \left( {\ln \frac{{{{\left| {\Delta {x_j}\left( t_s \right)} \right|}^{{{\lambda} _1} - {{\lambda} _2} - 1}}}}{{{{\left| {\Delta {x_j}\left( t_s \right)} \right|}^{{{\lambda} _1} - {{\lambda} _2} - 1}} + {\raise0.7ex\hbox{${{\alpha _2}}$} \!\mathord{\left/
											{\vphantom {{{\alpha _2}} {{\alpha _3}}}}\right.\kern-\nulldelimiterspace}
										\!\lower0.7ex\hbox{${{\alpha _3}}$}}}}} \right)}}}{{\left( {{{\lambda} _1} - {{\lambda} _2} - 1} \right)\left( {{\alpha _2} - {\alpha _3}} \right)}}
		\end{equation}
		And there is $|\Delta \bm{x}_j(t_s+{t_{{\text{r}}j1}})| = \sqrt{\Delta}$. 
		
		When $|\Delta \bm{x}_j| < \sqrt{\Delta}$ , there is
		${\bm{s}_j}{{ = }}  {\alpha _3}{\left| {\Delta {\bm{x}_j}} \right|^{{{\lambda} _1} - {{\lambda} _2} \cdot \tanh [{\lambda}_3(- \Delta)]}}\operatorname{sgn} \left( {\Delta {\bm{x}_j}} \right)+{\alpha _1}\Delta {\dot {\bm{x}}_j} + {\alpha _2}\Delta {\bm{x}_j}$.
		Similarly, it can be obtained that:
		\begin{equation}
			{t_{{\text{r}}j2}} =   \frac{{{\alpha _1 \left( {\ln \frac{{{{\left| {\Delta {x_j}\left( t_s+{t_{{\text{r}}j1}} \right)} \right|}^{{{\lambda} _1} - {{\lambda} _2}\cdot \tanh [{\lambda}_3(- \Delta)] - 1}}}}{{{{\left| {\Delta {x_j}\left( t_s+{t_{{\text{r}}j1}} \right)} \right|}^{{{\lambda} _1} - {{\lambda} _2}\cdot \tanh [{\lambda}_3(- \Delta)] - 1}} + {\raise0.7ex\hbox{${{\alpha _2}}$} \!\mathord{\left/{\vphantom {{{\alpha _2}} {{\alpha _3}}}}\right.\kern-\nulldelimiterspace}\!\lower0.7ex\hbox{${{\alpha _3}}$}}}}} \right)}}}{{\left( {{{\lambda} _1} - {{\lambda} _2}\cdot \tanh [{\lambda} _3(- \Delta)] - 1} \right)\left( {{\alpha _2} - {\alpha _3}} \right)}}.
		\end{equation}
		And there is $|\Delta \bm{x}_j(t_s+{t_{{\text{r}}j1}}+{t_{{\text{r}}j2}})| = 0$. 
	}
	\item {If $|\Delta \bm{x}_j(t_s)| < \sqrt{\Delta}$, there is
		${\bm{s}_j}{{ = }}  {\alpha _3}{\left| {\Delta {\bm{x}_j}} \right|^{{{\lambda} _1} - {{\lambda} _2} \cdot \tanh [{\lambda} _3(- \Delta)]}}\operatorname{sgn} \left( {\Delta {\bm{x}_j}} \right)+{\alpha _1}\Delta {\dot {\bm{x}}_j} + {\alpha _2}\Delta {\bm{x}_j}$.
		It can be obtained that:
		\begin{equation}
			{t_{{\text{r}}j1}} =  0 
		\end{equation} \vspace{-0.8cm}
		\begin{equation}
			{t_{{\text{r}}j2}} =   \frac{{{\alpha _1 \left( {\ln \frac{{{{\left| {\Delta {x_j}\left( t_s \right)} \right|}^{{{\lambda} _1} - {{\lambda} _2}\cdot \tanh [{\lambda} _3(- \Delta)] - 1}}}}{{{{\left| {\Delta {x_j}\left( t_s \right)} \right|}^{{{\lambda} _1} - {{\lambda} _2}\cdot \tanh [{\lambda} _3(- \Delta)] - 1}} + {\raise0.7ex\hbox{${{\alpha _2}}$} \!\mathord{\left/{\vphantom {{{\alpha _2}} {{\alpha _3}}}}\right.\kern-\nulldelimiterspace}\!\lower0.7ex\hbox{${{\alpha _3}}$}}}}} \right)}}}{{\left( {{{\lambda} _1} - {{\lambda} _2}\cdot \tanh [{\lambda} _3(- \Delta)] - 1} \right)\left( {{\alpha _2} - {\alpha _3}} \right)}}
		\end{equation}
		
		There is $|\Delta \bm{x}_j(t_s+{t_{{\text{r}}j1}}+{t_{{\text{r}}j2}})| = 0$. And the whole convergency time can be estimated roughly as $\mathop{\max}\limits_{{j}=1,2,3} \left( t_\mathrm{s}+{t_{{\text{r}}j1}}+{t_{{\text{r}}j2}}\right) $.
	}
\end{enumerate}

\section{Experimental Results}
\noindent In this section, a series of experiments are conducted to evaluate both convergency performance and robustness of the control algorithm proposed in this paper. Since the estimation accuracy of Jacobian matrix is important to the system’s performance, both the estimation error and convergency error will be evaluated so that characters of different estimation methods and controllers will be explored.

The simulation experiments are carried out in CoppeliaSim. The experiment involves a UR5 robot arm and a depth camera. And the depth camera is hanging above the robot arm. A red ball is fixed on the robot arm's end to make image identification easier. And the layout is shown in Fig.5.

\begin{figure}[htbp]
	\centering
	\includegraphics[width=8cm]{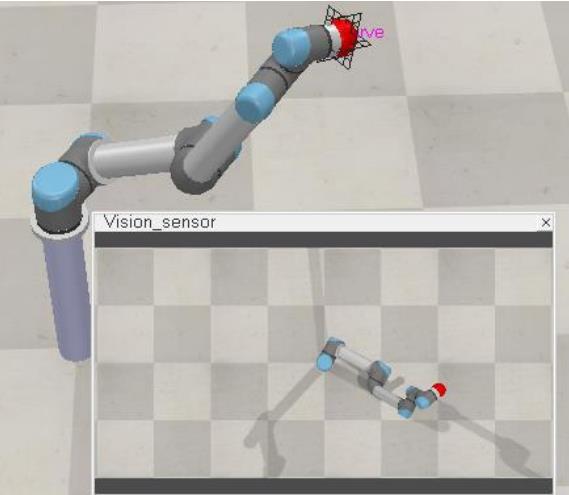}
	\caption{The overview of simulation environment in CoppeliaSim. The image in ‘Vision sensor’ window is the image captured by the depth camera.}
\end{figure}

\subsection{Validation of the Jacobian matrix estimation}

In this section, four Jacobian matrix estimators named “LKF”, “UKF”, “RLS”, and “Proposed” separately are taken into consideration, and their effectiveness will be assessed via two different criteria, which are shown in (40) and (41). The UR5 robot will conduct a smooth motion in joint space preset in advance. Besides, the parameter n1 of proposed estimator in (17) is set at zero because the motion is totally arranged in joint space and there is no destination.

\begin{equation}
	{T_1} = \left\| {\Delta {{\bm{s}}_k} - {{{\bm{\hat J}}}_k}\Delta {{\bm{r}}_k}} \right\|
\end{equation}  

\begin{equation}
	{T_2} = \left\| {{{\bm{s}}_k} - {{{\bm{\hat s}}}_k}} \right\|
\end{equation}  

where the criteria of ${T_2}$ represents the accumulation error of estimator and the sign ${{{\bm{\hat s}}}_k}$ is expressed as:
\begin{equation}
	{{\bm{\hat s}}_k} = {{\bm{\hat s}}_{k - 1}} + {{\bm{\hat J}}_k}\Delta {{\bm{r}}_k}
\end{equation}  
In this case, ${T_1}$ can reveal the magnitude of the accumulated error from the beginning of the motion to the present moment. The plot of criteria of ${T_1}$ and ${T_2}$ during the motion is demonstrated in Fig.6 and Fig.7 
\begin{figure}[htbp]
	\centering
	\includegraphics[width=10cm]{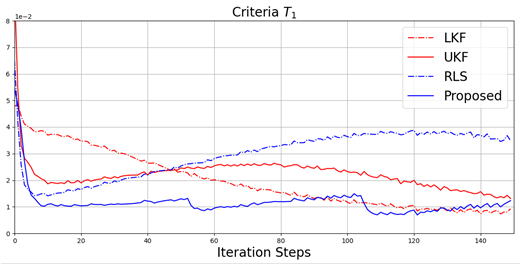}
	\caption{Profiles of the criteria $T_1$ that are computed along the trajectory in joint space.}
\end{figure}
\begin{figure}[htbp]
	\centering
	\includegraphics[width=10cm]{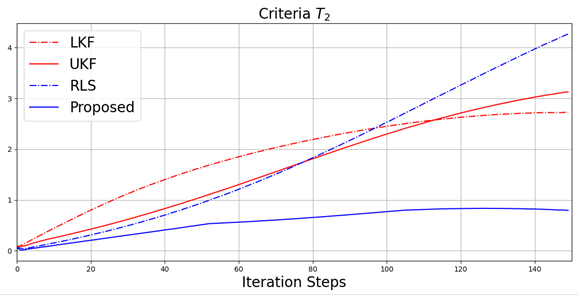}
	\caption{Profiles of the criteria $T_2$ that are computed along the trajectory in joint space.}
\end{figure}

It can be learnt from the ${T_1}$ curve that RLS estimator, UKF estimator, and Proposed estimator possesses a relatively accurate estimation at the beginning. However, RLS estimator can hardly suppress the estimation error in a small range compared with the other three. The LKF estimator can quickly reduce the estimation error which is relatively large though. The UKF estimator and proposed estimator can achieve an accurate initial estimation and keep the accuracy high during the motion. Among these four estimators, proposed one is an ideal choice, which balances the initial accuracy and the accuracy in the process because of the introduction of neural network and update algorithm. It can be learnt obviously from the ${T_2}$ curve that the proposed estimator is the best among four estimators. But there is still disadvantage of proposed method, the ability to correct the estimation error is related with estimation error and the larger previous error may trigger stronger amendment, which may cause an estimation error curve which is not that smooth. What’s more, the introduction of neural network can lead to a better initial estimation, it may also, however, yield an upper boundary on the accuracy. Both these disadvantages are demonstrated in ${T_1}$ plot. 
\subsection{Manipulation}
In this section, the UR5 robot is controlled by the velocity controller in order to drive the end-effector moving to the desired position. Other three control experiments are carried out in this section to demonstrate the effect of the algorithm proposed in this paper. One derive from \cite {17}, named RBF+PID, where the Jacobian matrix is estimated via RBFNN and the neural network is updated in control process. It is similar in some respects to the controller proposed in this paper and they will be based on the same neural network so the effect of different controller and update algorithm can be revealed. The other two controllers, separately called MPC and MFAC, use totally online method and are based on UKF estimator since it outperforms LKF estimator and RLS estimator in Section 4.1. And their detailed designs are as follows.

\begin{itemize}
	\item {Method1:RBF+PID scheme formula \cite {17}:\\
		In this scheme, the Jacobian matrix estimator is constructed by a RBFNN, which is the same as the estimator in our paper. Besides, the Lyapunov function used in \cite {17} consists of both system error and estimator error. Its controller, however, is designed merely based on the decrease of Lyapunov function and this is the main different between this scheme and proposed scheme. Its detailed design are as follows. The RBFNN estimator from (43) to (44) is the same as the estimator in proposed scheme but the update algorithm in (45) is different. And the controller is shown in (46).
		\begin{equation}
			{\bm{\theta} _{{{it}}}} = {{\rm{e}}^{{{ - }}\left[ \frac{{||\bm{r} - {\bm{u}_{{{it}}}}||}}{{{\bm{\delta} _{{{it}}}}}}\right] {^2}}}  = {{\rm{e}}^{{{ - }}\left[ \sqrt {\sum\limits_{{{j}} = 1}^{{l}} {{{(\frac{{{\bm{r}_{{j}}} - {\bm{u}_{{{itj}}}}}}{{{\bm{\delta} _{{{it}}}}}})}^2}} }\right]  {^2}}}
		\end{equation} \vspace{-0.8cm} 
		\begin{equation}
			{\hat {\bm{J}}_{{i}}} = {\bm{W}_{{i}}}{\bm{\theta} _{{i}}}(\bm{r}),{{i}} = 1,...,6
		\end{equation}  \vspace{-0.8cm}
		\begin{equation}
			{\dot {\hat {\bm{W}}}}^\mathrm{T}_{{{ij}}} = {\dot {\bm{r}}_{{i}}}{\bm{\theta} _{{i}}}({{{n}}_2}{{\bm{e}}_{{j}}} + {n_3}{\bm{s}_j})
		\end{equation}  \vspace{-0.8cm}
		\begin{equation}
			{\dot {\bm{r}} =  - }k{\left( {\hat {\bm{J}}} \right)^ + }\Delta \bm{x}
		\end{equation}     \vspace{-0.8cm}\\
		By deriving the Lyapunov function ${\rm{V}} = \frac{1}{2}{n_2}\Delta {\bm{x}^\mathrm{T}}\Delta \bm{x} + \frac{1}{2}\sum\limits_{{\rm{i}} = 1}^{\rm{n}} {\sum\limits_{{\rm{j}} = 1}^{\rm{l}} {\Delta {\bm{W}_{{\rm{ij}}}}\Delta \bm{W}_{{\rm{ij}}}^\mathrm{T}} } $, 
it can be obtained that:
\begin{equation}
	\begin{aligned}
		{\rm{\dot V}} =&{n_2}\Delta {\bm{x}^\mathrm{T}}\Delta \dot {\bm{x}} - \sum\limits_{{\rm{i}} = 1}^{\rm{n}} {\sum\limits_{{\rm{j}} = 1}^{\rm{l}} {\Delta {\bm{W}_{{\rm{ij}}}}{\dot {\hat {\bm{W}_{{\rm{ij}}}}}}^\mathrm{T}} } \\
		=& {n_2}\Delta {\bm{x}^\mathrm{T}}(\bm{e} + \hat {\bm{J}}\dot {\bm{r}}) - \sum\limits_{{\rm{i}} = 1}^{\rm{n}} {\sum\limits_{{\rm{j}} = 1}^{\rm{l}} {\Delta {\bm{W}_{{\rm{ij}}}}{{\dot {\bm{r}}}_{\rm{i}}}{\bm{\theta} _{\rm{i}}}({{\rm{n}}_2}\Delta {\bm{x}_{\rm{j}}} + {{\rm{n}}_3}{\bm{e}_{\rm{j}}})} } \\
		=&  - {\rm{k}}{n_2}\Delta {\bm{x}^\mathrm{T}}\Delta \bm{x} - {{\rm{n}}_3}{\bm{e}^\mathrm{T}}\bm{e} \le 0
	\end{aligned}
\end{equation}  
	}
	\item {Method2:UKF+MFAC scheme formula \cite {adaptive}:}\\
		In this scheme, the Jacobian matrix estimator is constructed via Unscented Kalman Filter (UKF), which possesses both ideal initial estimation and performance in control process. A model free adaptive controller (MFAC) is applied in this scheme. Similarly, the performance index is designed with the combination of system error and system input. By equating the derivative of performance index with respect to input to zero, the velocity command can be computed. Its detailed design are as follows and the detailed formulation of UKF estimator can be seen in \cite {adaptive}.\\
		The model free adaptive controller is designed as:
		\begin{equation}
			{r_k} = {r_{k - 1}} + {\left( {\lambda {{\bm{E}}_q} + {\bm{\hat J}}_k^{\rm{T}}{{{\bm{\hat J}}}_k}} \right)^{ - 1}}{\bm{\hat J}}_k^{\rm{T}}{e_{k - 1}}
		\end{equation}     \vspace{-0.8cm}	\\
		where $\lambda $ is the weight that controls the magnitude of $\Delta {r_k}$. Note that the model estimation algorithm (like UKF) exactly approximates the Jacobian matrix such that ${{\bm{J}}_k} = {{\bm{\hat J}}_k}$.	
	\item {Method3:UKF+MPC scheme formula \cite {towards}:}\\
		In this scheme, the Jacobian matrix estimator is constructed via UKF estimator. A model predictive controller (MPC) is applied in this scheme, where all states within the prediction horizon are taken into account. In the MPC controller, a criterion considering both smooth of manipulation and system error is designed, by setting its derivative with respect to input as zero, the velocity command can be derived. The detailed design of model predictive controller are as follows:
		\begin{equation}
			\begin{aligned}
					{r_k} =  & {r_{k - 1}} + {\left( {a{{{\bm{\hat J}}}_k} + {\bm{\hat J}}{{_k^{\rm{T}}}^ + }{\bm{Q}}} \right)^ + }(b - c){e_k}\\
					a =  & {{\left( {{H^2}{\alpha ^ * }^H - 2b} \right)} \mathord{\left/
							{\vphantom {{\left( {{H^2}{\alpha ^ * }^H - 2b} \right)} {\ln {\alpha ^ * }}}} \right.
							\kern-\nulldelimiterspace} {\ln {\alpha ^ * }}}\\
					b =  & {{\left( {H{\alpha ^ * }^H\ln {\alpha ^ * } - {\alpha ^ * }^H + 1} \right)} \mathord{\left/
							{\vphantom {{\left( {H{\alpha ^ * }^H\ln {\alpha ^ * } - {\alpha ^ * }^H + 1} \right)} {{{\ln }^2}{\alpha ^ * }}}} \right.
							\kern-\nulldelimiterspace} {{{\ln }^2}{\alpha ^ * }}}\\
					c =  & {{\left( {H{{\left( {{\alpha ^ * }\beta } \right)}^H}\ln \left( {{\alpha ^ * }\beta } \right) - {{\left( {{\alpha ^ * }\beta } \right)}^H} + 1} \right)} \mathord{\left/
							{\vphantom {{\left( {H{{\left( {{\alpha ^ * }\beta } \right)}^H}\ln \left( {{\alpha ^ * }\beta } \right) - {{\left( {{\alpha ^ * }\beta } \right)}^H} + 1} \right)} {{{\ln }^2}\left( {{\alpha ^ * }\beta } \right)}}} \right.
							\kern-\nulldelimiterspace} {{{\ln }^2}\left( {{\alpha ^ * }\beta } \right)}}
			\end{aligned}
		\end{equation}
	where $H$ represents the length of prediction horizon, $0 < \alpha^ *  \le 1$, $\beta  = \exp \left( { - \rho } \right)$, $\rho$ is a positive constant, and ${\bm{Q}}$ is a symmetric and positive definite matrix used to adjust the command.	
\end{itemize}

We carried out four experiments with a variety of different initial and desired position. The experiments and result are shown respectively in Fig.8. The convergency plots show that proposed scheme aimed at finite-time control considerably boost the convergency speed. Another trait of proposed scheme is that the three elements of feedback information seem to converge in a sequential order, which may cause fluctuations when an element researches the desired range in first, since the system is not decoupled. What’s more, the using of neural network when estimating Jacobian matrix may affect the performance of system if not well-trained (shown in Fig.8a and Fig.8d). 

During our experiments, we found some fault conditions. When the initial position is too far from desired position, the robot may encounter collision since the route planning is not considered here. In this case it is important to make sure that the distance from initial position to desired position should not be too large in practical experiments.

\begin{figure}[H]
	\centering
	\subfloat[]{
		\includegraphics[width=1.95 in]{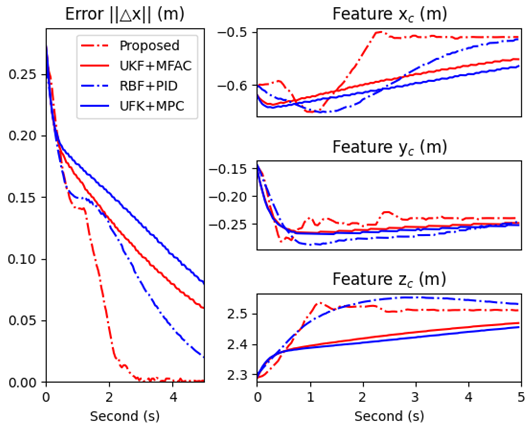}}
	\subfloat[]{
		\includegraphics[width=1.95 in]{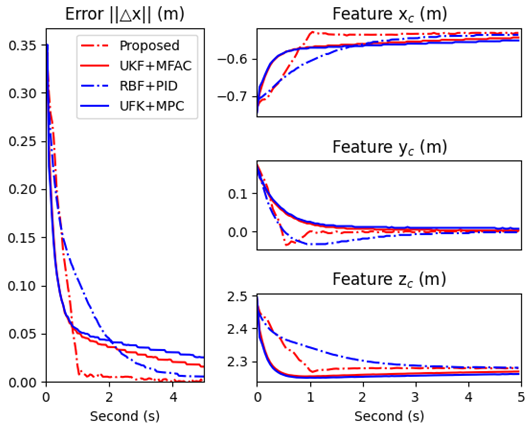}}\vspace{-0.3cm}
	\quad
	\subfloat[]{
		\includegraphics[width=1.95 in]{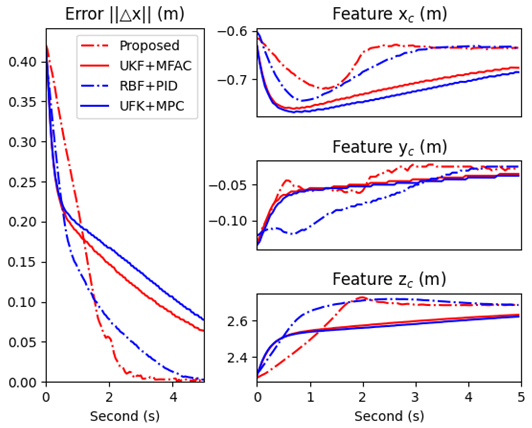}}
	\subfloat[]{
		\includegraphics[width=1.95 in]{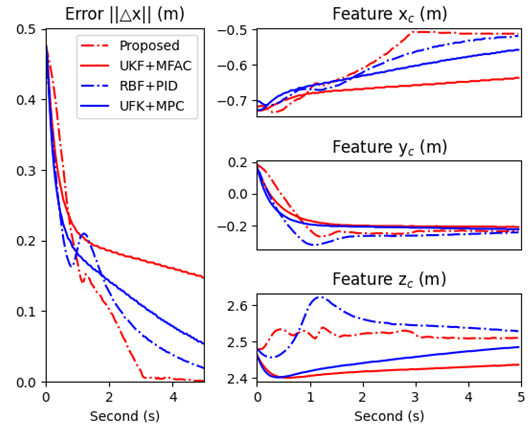}}\vspace{-0.3cm}
	\caption{Among four methods, the profiles of error of the four visual servoing experiments and value of image feature captured by depth camera. (a) EXP1. (b) EXP2. (c) EXP3. (d) EXP4 .}
\end{figure}

There are totally four schemes used in experiments. Among them, “UKF+MFAC” and “UKF+MPC” have independent estimators and their controllers are based on the assumption that estimation is reliable. And the stability proof of these two schemes either considers only system error or is not given. Both of the other two schemes consider the convergency of error of output and estimation in stability proof and the performance of these two schemes seems better in system error converging roughly. The MPC controller has similar performance as MFAC controller. Since their estimators are totally based on system input-output information and their controllers do not expect special control effect. The proposed scheme outperforms the RBF+UKF scheme in system error converging because of the introduction of finite-time control designing. It is apparent in Fig.8 that the proposed controller makes the error converge at an almost constant rate until the error converge to a small value. The overshoot of the 3D output value of the proposed scheme is almost the same as that of the “RBF+PID” scheme, but the proposed scheme has faster response speed and the convergence speed of the proposed scheme 3D output value is almost constant.

To show the trajectory of the end-effector of the manipulator for different controller, Fig.9 and Fig.10 demonstrate the trajectory of the end-effector of the manipulator from the depth camera perspective and its corresponding 3-dimensional plot, respectively.
\begin{figure}[H]
	\centering
	\subfloat[]{
		\includegraphics[width=3.2cm]{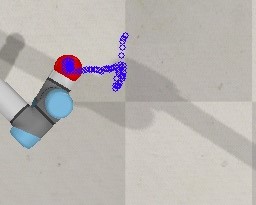}}
	\subfloat[]{
		\includegraphics[width=3.2cm]{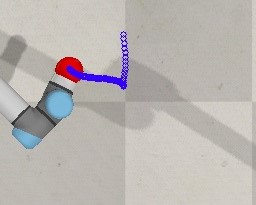}}
	\subfloat[]{
		\includegraphics[width=3.2cm]{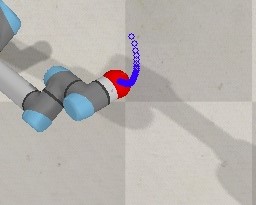}}
	\subfloat[]{
		\includegraphics[width=3.2cm]{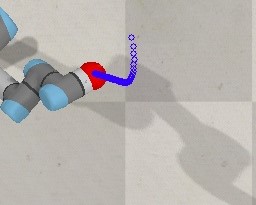}} \vspace{-0.3cm}
	\quad 
	\subfloat[]{
		\includegraphics[width=3.2cm]{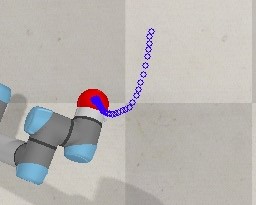}}
	\subfloat[]{
		\includegraphics[width=3.2cm]{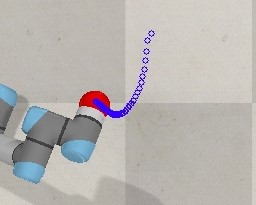}}
	\subfloat[]{
		\includegraphics[width=3.2cm]{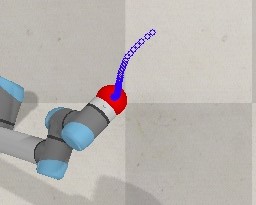}}
	\subfloat[]{
		\includegraphics[width=3.2cm]{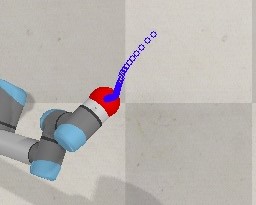}} \vspace{-0.3cm}
	\quad
	\subfloat[]{
		\includegraphics[width=3.2cm]{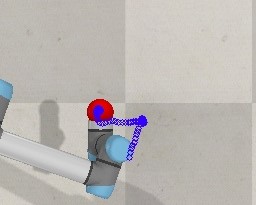}}
	\subfloat[]{
		\includegraphics[width=3.2cm]{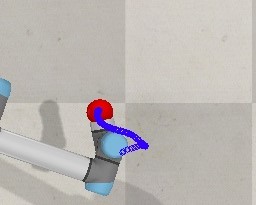}}
	\subfloat[]{
		\includegraphics[width=3.2cm]{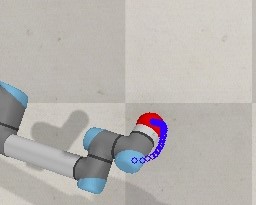}}
	\subfloat[]{
		\includegraphics[width=3.2cm]{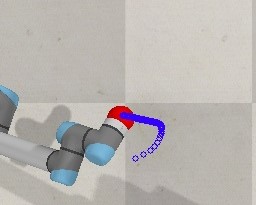}} \vspace{-0.3cm}
	\quad
	\subfloat[]{
		\includegraphics[width=3.2cm]{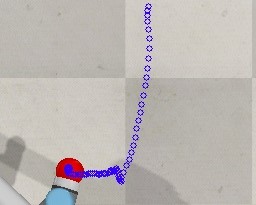}}
	\subfloat[]{
		\includegraphics[width=3.2cm]{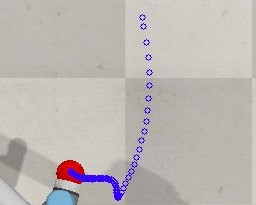}}
	\subfloat[]{
		\includegraphics[width=3.2cm]{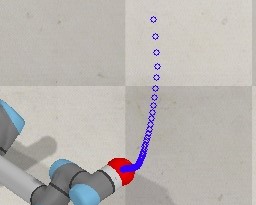}}
	\subfloat[]{
		\includegraphics[width=3.2cm]{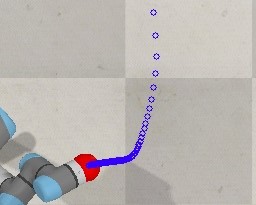}} \vspace{-0.3cm}
	\caption{The end trajectory in the four visual servoing experiments which have a variety of different start and target positions with four control algorithms, and the current position is at the target position. Each circle corresponds to the end-effector position of a control interval time. (a) EXP1:Proposed. (b) EXP1:RBF+PID. (c) EXP1:UKF+MFAC. (d) EXP1:UKF+MPC. (e) EXP2:Proposed. (f) EXP2:RBF+PID. (g) EXP2:UKF+MFAC. (h) EXP2:UKF+MPC. (i) EXP3:Proposed. (j) EXP3:RBF+PID. (k) EXP3:UKF+MFAC. (l) EXP3:UKF+MPC. (m) EXP4:Proposed. (n) EXP4:RBF+PID. (o) EXP4:UKF+MFAC. (p) EXP4:UKF+MPC.}
	\label{fig.1}
\end{figure}     \vspace{-0.4cm}

\begin{figure}[H]
	\centering
	\subfloat[]{ 
		\includegraphics[width=1.95 in]{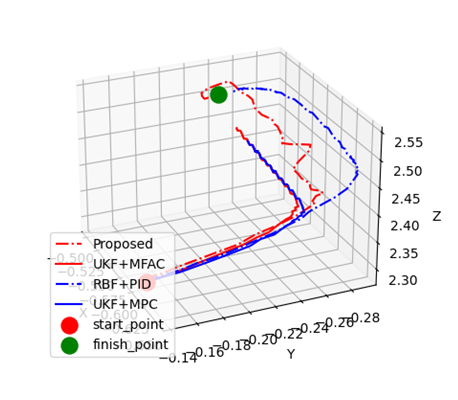}}
	\subfloat[]{
		\includegraphics[width=1.95 in]{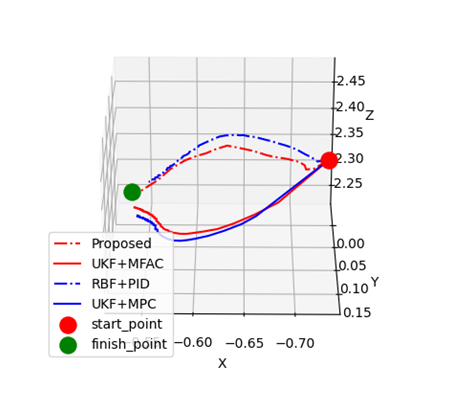}} \vspace{-0.3cm}
	\quad
	\subfloat[]{
		\includegraphics[width=1.95 in]{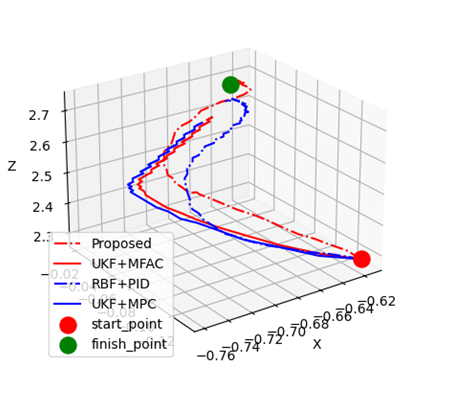}}
	\subfloat[]{
		\includegraphics[width=1.95 in]{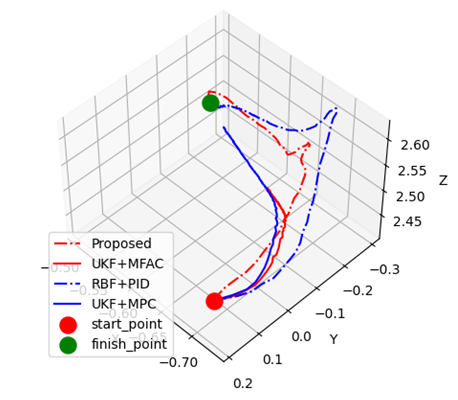}} \vspace{-0.3cm}
	\caption{The 3-dimensional plot of the end trajectory in the four visual servoing experiments which have a variety of different start and target positions with four control algorithms, and the current position is at the target position. (a) EXP1. (b) EXP2. (c) EXP3. (d) EXP4 .}
\end{figure}

\section{Conclusion}
\noindent In this paper, we propose a finite-time control algorithm in a combination of offline and online way, where the estimation of local hand-eye relationship is carried out via an adaptive neural network. Due to the combination of online and offline method, the proposed estimator can fetch a balance between performance and robustness which outperforms the estimators that only use online method. Moreover, the introduction of a finite-time control also boosts the convergency speed of proposed algorithm, and specific working time can be roughly evaluated. The adaptive factor introduced in controller designing helps the system converge in fast speed when system error is large and helps the system keep stable when the system error is small. Several experiments are carried out to illustrate the characteristics of proposed algorithm. Via experiments and numerical tool, the effectiveness and reliability of proposed method are validated.

SGPFS method helps system state move to the sliding surface in finite time, which is realized by finite-time controller and weights matrix of RBFNN. Moreover, the designing of fast terminal sliding mode surface makes the state quickly converge to the equilibrium when moving on the sliding surface. In summary, this paper proposed a feasible method for finite-time control and estimation task for visual servoing.

However, the proposed method also has some limitations. Firstly, the method only concerns the end position of manipulator. Secondly, the offline training of RBFNN requires a large amount of dataset to train itself, which may become a roadblock for its application.

\section*{Acknowledgements}
\noindent Haibin Zeng makes the conception of the study, writes the manuscript, and carries out the experiment;
Yueyong Lyu helps the study with constructive discussions; 
Jiaming Qi and Shuangquan Zoua provided stimulating discussions;
Tanghao Qin and Wenyu Qin contribute significantly to manuscript preparation.

\section*{Disclosure statement}
\noindent No potential conflict of interest was reported by the authors.

\section*{Funding}
\noindent This work was supported in part by National Key Research and Development Program of China, 2020YFB1506700, and National Natural Science Foundation (NNSF) of China under Grant 12150008, 61973100, 61876050. 

\section*{Notes on contributors}
\noindent Haibin Zeng received the B.S. degree in Automation from Wuhan University of Science and
Technology, Wuhan, China, in 2021. He is currently working toward the M.S. degree in control
Engineering with the Harbin Institute of Technology, Harbin, China. His current research interests include robotic manipulation, and visual servoing.

\noindent Yueyong Lyu received the bachelor’s, master’s,
and Ph.D. degrees from the Harbin Institute of
Technology, Harbin, China, in 2002, 2008 and
2013, respectively. He is currently an Associate Research Fellow
with the Department of Control Science and Engineering, Harbin Institute of Technology, Harbin,
China. His interests of research mainly focus
on spacecraft guidance, navigation and control,
especially in spacecraft formation flying, on-orbit
service, etc.

\noindent Jiaming Qi received the M.Sc. degree in integrated circuit engineering from the Harbin Institute of Technology, Harbin, China, in 2018. He is currently working toward the Ph.D. degree
in Control Science and Engineering with Harbin
Institute of Technology, Harbin, China.
In 2019, he was a visiting Ph.D. student with
The Hong Kong Polytechnic University. His current research interests include robotic manipulation, data-driven control, and visual servoing.

\noindent Shuangquan Zou received the M.Sc. in control
Engineering from Harbin Institute of Technology in 2020. He is currently pursuing the Ph.D.
degree with Control Science and Engineering,
Harbin Institute of Technology. His current research interests include shape detection for soft
manipulator, kinematics model, and machine vision

\noindent Tanghao Qin received the M.Sc. degree in control Engineering from Harbin Institute of Technology in 2022. He is currently pursuing the Ph.D. degree in Aeronautical and Astronautical
Science and Technology with Harbin Institute
of Technology. His current research interests
include robotic manipulation and formation control.

\noindent Wenyu Qin received the M.Sc. degree in control
Engineering from Harbin Institute of Technology
in 2022. He is currently pursuing the Ph.D. degree in Control Science and Engineering with
Harbin Institute of Technology. His current research interests include flexible spacecraft and
multi-agent systems.

\bibliographystyle{tfnlm}
\bibliography{ref}

\end{document}